\pdfoutput=1

\documentclass[11pt,table]{article}

\usepackage[final]{acl}

\usepackage[utf8]{inputenc}
\usepackage[T1]{fontenc}

\usepackage{microtype}

\usepackage{inconsolata}

\usepackage{graphicx}
\usepackage{booktabs,multirow}

\title{eFontes. Part of Speech Tagging and Lemmatization of Medieval Latin Texts.\\A Cross-Genre Survey}

\author{
	Krzysztof Nowak \\ 
	Institute of Polish Language (Polish Academy of Sciences) \\ \texttt{krzysztof.nowak@ijp.pan.pl} \\\AND
	Jędrzej Ziębura \\ 
	AGH University / Enelpol \\
	\texttt{ziebura.jedrzej@gmail.com} \\\AND
	Krzysztof Wróbel \\ 
	Jagiellonian University / Enelpol \\
	\texttt{krzysztof@wrobel.pro} \\\AND
	Aleksander Smywiński-Pohl \\ 
	AGH University / Enelpol \\  \texttt{apohllo@o2.pl} }

\begin{document}
	\maketitle
	
	\begin{abstract}
	This study introduces the eFontes models for automatic linguistic annotation of Medieval Latin texts, focusing on lemmatization, part-of-speech tagging, and morphological feature determination. Using the Transformers library, these models were trained on Universal Dependencies (UD) corpora and the newly developed eFontes corpus of Polish Medieval Latin. The research evaluates the models' performance, addressing challenges such as orthographic variations and the integration of Latinized vernacular terms. The models achieved high accuracy rates: lemmatization at 92.60\%, part-of-speech tagging at 83.29\%, and morphological feature determination at 88.57\%. The findings underscore the importance of high-quality annotated corpora and propose future enhancements, including extending the models to Named Entity Recognition.
	\end{abstract}	
	
	\section{Introduction}
	\label{sec:intro}
	After the decline of the Roman Empire, Latin continued to be widely utilized throughout Europe for over ten centuries. Latin served as the primary mode of communication across diverse social and cultural contexts, ranging from church and secular administrative records to scholarly writings in burgeoning universities, historical narratives, literary works, religious poetry, and liturgical texts. Given the critical role of Latin writings in understanding European culture and history during the Middle Ages, their sheer volume necessitates the application of distant reading techniques for their analysis and better comprehension.

Developing a pipeline for the automatic processing of Latin texts is far from straightforward due to the language's diverse applications and extensive history. In recent years, numerous studies have emerged focusing on the development of methods for processing historical languages, Latin. However, the majority of these studies have concentrated on Classical and Late Latin. (see section \ref{sec:previous} below).

The specificity of the Medieval Latin \citep{stotzHandbuchZurLateinischen1996a} may pose substantial challenges for processing tools. Beyond its vast range of uses, this is mainly due to its pan-European nature as a \textit{lingua franca} throughout almost all of medieval Europe. The standardization of Latin across a region spanning from Sweden to Italy and from Poland to Portugal occurred as the language was disseminated through education and utilized predominantly in formal contexts. Despite the overarching influence of the church or secular authorities in unifying the language, significant regional variations also emerged. Additionally, Latin interacted closely with vernacular languages, impacting many aspects of their pronunciation, grammar, vocabulary and syntax. Furthermore, Medieval Latin frequently was intertwined with vernacular languages in manuscripts. Consequently, the documents and court records produced during the Middle Ages often contain a significant number of words in medieval Polish, German, or English. \citep{goyensDawnWrittenVernacular2003}

This paper introduces a family of \textit{eFontes} models for automatic annotation of the Medieval Latin texts. Based on the Transformers library \cite{wolf2020transformers}, they  include models for context-independent lemmatization, part of speech and morphological features tagging. The models were trained on a set of publicly available Universal Dependencies corpora (henceforth: UD corpora) and a new language resource, namely the corpus \textit{eFontes} of Polish Medieval Latin. In section \ref{sec:previous}, we briefly summarize the research on automatic annotation of Latin language texts. In sections \ref{sec:lrs} and \ref{sec:system}, we present our system and the datasets used in training and evaluation. In section \ref{sec:results} we evaluate the performance of the models and provide a thorough error analysis for the Lemmatization and PoS tagging tasks. In section \ref{sec:conclusions}, we provide a concise summary of the results, propose areas for improving the models, and outline plans for extending the models for the Named Entity Recognition.
	
	\section{Previous Work}
	\label{sec:previous}
	In their recent survey, \citet{sommerschieldMachineLearningAncient2023} observe that the research on the automatic processing of ancient languages has significantly accelerated in recent years. Out-of-the-box models for user-friendly processing of Latin texts are integrated into popular frameworks like \textit{Classical Language Toolkit} \citep{johnson-etal-2021-classical} and SpaCy. The latter now offers several packages with pre-trained models for Latin, such as spacy-udpipe\footnote{\url{https://spacy.io/universe/project/spacy-udpipe}} and LatinCy\footnote{\url{https://spacy.io/universe/project/latincy}}

As far as custom solutions are concerned, an important stimulus and a platform for presenting progress in the field turned out to be the biannual workshops LT4HALA. Language Technology for Historical and Ancient Languages \citep{sprugnoliOverviewEvaLatin20202020,sprugnoliOverviewEvaLatin20222022}. At the EvaLatin competition hosted during the 2022 edition of the workshop, \citet{WROBELNOWAK} presented a \textsc{Cracovia} system that outperformed other contributions in every tagging task for both cross-genre and cross-time corpus. Although the present solution is based on this work, there is no shortage of new systems.

Recently, \citet{riemenschneiderExploringLargeLanguage2023} conducted a thorough evaluation of existing systems across various natural language processing tasks, such as part-of-speech tagging and lemmatization. Their custom models for Latin outperformed the \textsc{Cracovia} tagger, except for the cross-genre and cross-time subtasks, where the open modality system of \citet{WROBELNOWAK} exhibited better performance.

The challenge of applying tools built for Classical Latin in tagging medieval texts was recognized early on. In the \textsc{Omnia} project, for example, TreeTagger was trained on a manually annotated corpus of medieval Latin on top of previously available Classical Latin parameters \citep{bonOMNIAOutilsMethodes2011}.

\citet{eger-etal-2016-lemmatization} compared the performance of the solutions existing at the time, in lemmatization and part-of-speech tagging of medieval capitularies, and showed how the performance of the system may be enhanced by incorporating word embeddings and lexicon rules into the picture. Their study revealed a notable decrease in accuracy (below 90\%) when a tagger trained on Classical or Late Latin corpus was applied to medieval texts. Building upon this research, \citet{kestemontIntegratedSequenceTagging2016} introduced a joint learning solution for the integrated approach to lemmatization and part-of-speech tagging. Their study showed, among other insights, the impact of normalizing orthographic variations in medieval texts on the accuracy of the system.

Importantly, significant advancements have also been made in processing medieval variations of Romance languages, such as Old French \cite{campsCorpusModelsLemmatisation2021}.

	\section{Latin Datasets Used}
	\label{sec:lrs}
	As demonstrated in section \ref{sec:system} below, the models presented in this paper were trained and evaluated using a set of publicly available UD corpora and the \textit{eFontes} corpus of Polish Medieval Latin. Since we aimed to assess the impact of using data from different periods and areas of Latin language use, we provide a more detailed presentation of the corpora in the following section, focusing on the features relevant to this study.

\subsection{UD Corpora}
\label{sec:ud.corpora}

The UD-conformant corpora utilized in training and evalution of the models include:

\begin{itemize}
    \item The \textbf{Perseus} corpus which contains works composed between the 1st century BCE and the 2nd century CE with the exception of the Jerome's translation of the Vulgate from the 4th--5th century BC. The dataset includes fragments of a historical work by Tacitus, Cicero's speeches by Cicero, Phaedrus' \textit{Fables} and poetical works of Propertius.
    
    \item The \textbf{PROIEL} corpus \citep{Haug2008CreatingAP} consists of selected books of the Jerome's \textit{Vulgate}, and selected fragments of the historical work of Caesar, Cicero's letters, as well as scholarly treatises by Palladius (on agriculture from the 4th century CE) and Cicero (\textit{de officiis} written in the 1st century BCE).
    
    \item The \textbf{LLCT} (\textit{Late Latin Charter Treebank}) \citep{cecchini-etal-2020-new} is a large collection of medieval charters written between the 8th and the 9th century in Tuscany.
    
    \item The \textbf{ITTB} (\textit{Index Thomisticus Treebank}) consists of the annotated works of Thomas Aquinas written in the 13th century \citep{lait-ud-2018}.
    
    \item Finally, the \textbf{UDante} Treebank \citep{cecchiniUDanteFirstSteps2020} includes both prosaic (treatises on linguistics, politics, and a selection of letters) and poetical (\textit{Eclogues}) works of Dante Alighieri.
\end{itemize}

\begin{table*}[htbp]
\scriptsize
\begin{center}
\begin{tabular}{>{\textsc}l ccc cccc}
\toprule
Corpus & \multicolumn{3}{c}{Coverage}  & \multicolumn{3}{c}{Number of} \\
\cmidrule(lr){2-4} \cmidrule(lr){5-7}
 & time (centuries) & place & text type & tokens & sentences & avg\\
\midrule
PROIEL & 1 BCE - 5 CE & Roman Empire & various & 177 558 & 16 196 & 10.96\\
Perseus & 1 - 5 CE & Roman Empire & various & 18 425 & 1 334 & 13.81\\
LLCT & 8 - 9 CE & Italy & charters & 390 819 & 7 289 & 26.64\\
ITTB & 13 CE & Italy & theological treatises & 390 819 & 22 775 & 17.16\\
UDante & 14 CE& Italy & various & 30 566 & 926 & 33.01\\
\bottomrule
\end{tabular}
\caption{Number of tokens, sentences and average number of tokens in a sentence in Universal Dependencies corpora used in the scenarios involving UD data.}
\end{center}
\end{table*}
\label{tab:ud_data}

Regarding their linguistic features (see Table \ref{tab:ud_data}), the Perseus and the PROIEL corpora represent mainly Classical, Post-Classical and Late Latin, whereas the other three corpora consist of texts written during the Middle Ages. The Perseus and PROIEL corpora exhibit relatively heterogeneous content, encompassing both prose and poetry across various genres (e.g., speeches, treatises, Bible translations) and covering a wide array of topics ranging from history to philosophy to agriculture. The LLCT treebank exclusively focuses on one genre, namely charters, while both UDante and ITTB contain texts attributed to a single author. The pre-medieval texts were composed within the boundaries of the Roman Empire, whereas medieval charters and the works by Thomas Aquinas and Dante were penned in what is now Italy.

Therefore, at first glance, the language represented in the UD treebanks appears to be relatively different from that of Polish Medieval Latin texts from the representative \textit{eFontes} corpus, even if we acknowledge that medieval Latin tended to be more conservative and less prone to change due to the stabilizing impact of writing.

\subsection{The \textit{eFontes} Corpus}
The \textit{eFontes} corpus has been compiled since 2013 and is expected to contain over 15 million tokens by the mid-2024. It comprises texts composed between 1000 and 1550 on the territory of the Kingdom of Poland. The corpus's representativeness is carefully monitored with regard to time, place, and text types.

The corpus is planned to be further expanded in the coming years, facilitated by new critical editions and the broader adoption of Handwritten Text Recognition technology. This, coupled with its potential for linguistic and historical research, underscores the importance of automatic annotation.

\begin{table}[htbp]
\begin{center}
\begin{tabular}{@{}lrrr@{}}

      \toprule
      Genre&Tokens&Sentences&Avg\\
      \midrule
      Annals & 895 & 33 & 27.12\\
     Biography & 8994 & 298 & 30.18\\
     Normative & 3142 & 115 & 27.32\\
      Proceedings & 7189 & 389 & 16.48\\
      Science & 1990 & 106 & 18.74\\          
      \bottomrule

\end{tabular}
\caption{Number of tokens, sentences and average number of tokens in sentence in data used in cross-validation.}
 \end{center}
\end{table}
\label{tab:efontes_data}

For training and evaluation purposes, a small manually annotated gold corpus was prepared based on texts from the \textit{eFontes} corpus. The composition of the dataset reflects most prominent text types and its domain and register variation (see Table \ref{tab:efontes_data}). The gold corpus comprises following genres:

\begin{itemize}
\item \textbf{Annals}
The genre is represented by the \textit{Annals of the Cistercian Order of Henrykow} from the 13/14 century.

\item \textbf{Biography} The subcorpus includes samples of popular hagiographic works:
    \begin{itemize}
    \item the \textit{Life of Anne, Duchess of Silesia} (Lat. \textit{Vita Annae ducissae Silesiae}) from the second half of the 13th century,
    \item the \textit{Miracles of Saint Adalbert} (Lat. \textit{Miracula Sancti Adalberti}) from the end of the 13th century, and
    \item the \textit{Life of Saint Kinga} (Lat. \textit{Vita Sanctae Kyngae}) from the first half of the 14th century.
    \end{itemize}
    
\item \textbf{Normative} The group consists of statutory texts concerning ecclesiastical law, in particular in includes so called synodial statutes of Gniezno and Kraków from the beginning of the 15th century.    
\item \textbf{Science} Scientific writing is represented by Vitello's technical treatise on optics completed by the end of the 13th century (Lat. \textit{Perspectivae liber primus});

\item \textbf{Proceedings}
The sub-corpus comprises a selection of records of courts of law and city councils and includes:
\begin{itemize}
        \item the book of the city of Lviv from the end of the 14th century;
        \item the books of the court of Kraków from the end of the 14th century;
        \item the book of a small village in the Southern Poland from the second half of the 15th century.
    \end{itemize}
\end{itemize}

The text samples included in the gold corpus were manually annotated by two highly-qualified philologists with expertise in medieval Latin linguistics and history. The part of speech, lemma and morphosyntactic features were annotated based on guidelines which followed the UD model.\footnote{The guidelines are to be published at \url{https://scriptores.pl/efontes}.} As mentioned earlier, the dataset was designed to reflect the diatopic, diachronic, and diastratic variation of Latin written production in Poland. It includes texts with features that pose systematic challenges for automatic processing tools, such as a large number of Latinized or non-Latinized Polish personal names, scientific and legal terminology, numerous medieval place names, and vernacular insertions ranging from single words to multi-word phrases. Moreover, the orthography of the texts is anything but consistent, with variations stemming from medieval scribal practices and modern editorial policies.
	
	\section{System Description}
	\label{sec:system}
	\subsection{Training scenarios}

In order to assess the influence of different training data on the results and specifically the necessity to use a custom Medieval Latin corpus we have designed several training scenarios.

In the first scenario (\textit{baseline}), the foundation models are fine-tuned only on the data from the \textit{eFontes} corpus. The training procedure follows a cross-validation scheme, where each \textit{eFontes} subcorpus is treated as a testing set and the remaining sub-corpora are used for training and validation.
Since the tasks are different and there are 5 subcorpora, the procedure yields 15 separate models, which are evaluated on the specific test sets.

In the second scenario (\textit{UD all}), the foundation models are fine-tuned on all the data from the UD Latin corpora (no normalization of spelling is applied) and tested for all tasks on all the sub-corpora. This scenario yields 3 models: one for each task. The scenario is designed to answer whether it is necessary to fine-tune the model on the \textit{eFontes} corpus, or if training solely on previously available data would be sufficient.

The third scenario involves using all the UD Latin corpora in the initial fine-tuning step. Subsequently, the model undergoes further fine-tuning on a specific UD subcorpus only (referred to as \textit{UD + specific UD corpus name}). With 5 available UD corpora and 3 tasks, this scenario yields 15 models, which are then evaluated on the individual \textit{eFontes} sub-corpora. This results in 75 evaluation outcomes. The scenario was designed to determine whether and which of the UD corpora exhibit the greatest similarities with the \textit{eFontes} domain-based sub-corpora, thereby making them more valuable for a specific set of documents.

The last scenario (\textit{UD + eFontes}) involves using all the data from the UD Latin corpora in the initial fine-tuning step. Then, similar to the \textit{baseline} scenario, the model undergoes further fine-tuning. The key difference between the \textit{baseline} scenario and this one is that the previous one uses the original foundation model, while this scenario employs a model that has already been fine-tuned on the UD Latin dataset. This scenario is designed to determine the impact of additional training data on the linguistic analysis tasks. Similar to the baseline scenario, it yields 15 individual models.

\subsection{Model architecture}

The architecture of all trained models is based on the transformer, as this family of models yield state-of-the-art results in part-of-speech tagging,
morphological feature determination and lemmatization \cite{van2021trankit}. 
The work builds on a morphosyntactic tagger KFTT \cite{publ252933} which won the PolEval 2020 task 1 contest (Morphosyntactic tagging of Middle, New and Modern Polish) as well as on the results presented during the EvaLatin 2022 competition by \citet{WROBELNOWAK}.

\subsubsection{POS and Morphological Features Tagging}

Part-of-speech and morphological features tagging tasks are addressed with a transformer encoder-only model with a \textit{token classification} head on top. 
\begin{table}[htb]
    \centering
    \begin{tabular}{l|l}
    \textbf{batch size} & 12 \\
    \textbf{epochs} & 10 \\
    \textbf{learning rate} & 2e-5 \\
    \textbf{sequence length} & 256 \\
    \end{tabular}
    \caption{The training parameters for Part of Speech Tagging and Morphological Features Determination tasks.}
    \label{tab:pos}
\end{table}

First, the transformer returns contextual embedding of each token, then a linear layer with a softmax activation returns normalized scores for each tag present in the training corpus.
During training both the pre-trained model and the classification head are updated, so the training uses a full back propagation procedure.
During pre-training XLM-R uses only masked language modelling (MLM) as a training task, so the model initially has no knowledge regarding the possible parts of speech or morphological tags (beyond knowledge that is extracted from the MLM task itself). For the baseline model which utilized only the \textit{eFontes} corpus, the classification head is initialized randomly. For all the remaining scenarios the model is already fine-tuned for the same task, but using a different corpus as described in the previous section. The important parameters of the training are given in Table \ref{tab:pos}. The remaining parameters were the defaults from the Huggingface library.

\begin{table}[htb]
    \centering
    \begin{tabular}{l|l}
     \textbf{batch size} & 128 \\
     \textbf{epochs} & 5 \\
     \textbf{input sequence length} & 48 \\
     \textbf{output sequence length} & 24 \\
     \textbf{learning rate} & 0.001 \\
    \end{tabular}
    \caption{The training parameters for Lemmatization tasks.}
    \label{tab:lemmatization}
\end{table}

\subsubsection{Lemmatization}
Lemmatization is a different task, with respect to the model architecture -- in general it requires from the model to produce an arbitrary sequence
of letters, depending on the input given. Taking that into account we have used ByT5 small model \citep{10.1162/tacl_a_00461} whose input are separate bytes of a text.
It is a text-to-text (or bytes-to-bytes) model, so it is well-suited for the task. We have performed some additional experiments with sub-word 
models such as mT5 \cite{xue-etal-2021-mt5}, but they clearly showed inferior performance.

The model receives as input the individual word to be lemmatized together with the predicted Part of Speech. The input is framed as the
inflected word and the PoS tag separated by a colon, e.g. \textit{adducam:VERB}.
The model does not receive the morphological features nor any information about the context of the lemmatized word.
The context is only indirectly reflected in the PoS tag provided.

The important parameters of the model training are given in Table \ref{tab:lemmatization}. The remaining parameters were the defaults from the Huggingface library.\footnote{The best models obtained in our experiment will be published at \url{https://huggingface.co/efontes}}



	\section{Results}
	\label{sec:results}
	\subsection{System Performance}
\begin{table*}[!htb]
\centering
\scriptsize
\ContinuedFloat
\begin{tabular}{lcccccc}
\toprule
 & \multicolumn{3}{c}{Biography} & \multicolumn{3}{c}{Normative} \\
 & UPOS & UFeats & Lemma & UPOS & UFeats & Lemma \\
\midrule
baseline & 95.43 & 82.06 & 80.75 & 95.81 & 81.53 & 85.05\\
UD all & 90.19 & 52.85 & 86.14 & 92.04 & 50.32 & 85.81 \\
UD + ITTB & 89.58 & \color{black} {\cellcolor[HTML]{EF553B}} 32.02 & 87.48 & 91.66 & 67.96 & 85.85 \\
UD + LLCT & 89.87 & 69.84 & 86.65 & 91.66 & 67.96 & 86.26 \\
UD + Perseus & 90.34 & 72.68 & \color{black} {\cellcolor[HTML]{EF553B}} 84.42 & 92.01 & 69.87 & \color{black} {\cellcolor[HTML]{EF553B}} 83.74 \\
UD + PROIEL & \color{black} {\cellcolor[HTML]{EF553B}} 77.34 & 75.78 & 85.10 & \color{black} {\cellcolor[HTML]{EF553B}} 79.46 & 71.95 & 84.73 \\
UD + UDante & 90.20 & 33.56 & 86.30 & 91.69 & \color{black} {\cellcolor[HTML]{EF553B}} 34.44 & 85.08 \\
UD + eFontes & \color{black} {\cellcolor[HTML]{B6E880}} 96.10 & \color{black} {\cellcolor[HTML]{B6E880}} 84.86 & \color{black} {\cellcolor[HTML]{B6E880}} 88.37 \\
\hline
\end{tabular}
\ContinuedFloat
\begin{tabular}{lccccccccc}
	\toprule
	& \multicolumn{3}{c}{Proceedings} & \multicolumn{3}{c}{Science} & \multicolumn{3}{c}{Annals} \\
	& UPOS & UFeats & Lemma & UPOS & UFeats & Lemma & UPOS & UFeats & Lemma \\
	\midrule
	baseline & 91.63 & 86.67 & 81.69 & 79.21 & 77.48 & 80.53 & 95.98 & 80.78 & \color{black} {\cellcolor[HTML]{B6E880}} 87.04 \\
	UD all & 93.18 & 67.12 & 79.82 & 75.35 & 51.88 & \color{black} {\cellcolor[HTML]{B6E880}} 96.45 & 86.15 & 59.11 & 82.79 \\
	UD + ITTB & 91.06 & 70.62 & 79.03 & 74.65 & \color{black} {\cellcolor[HTML]{EF553B}} 48.78 & 95.44 & 87.15 & \color{black} {\cellcolor[HTML]{EF553B}} 31.40 & 85.14 \\
	UD + LLCT & 91.06 & 70.62 & \color{black} {\cellcolor[HTML]{EF553B}} 78.95 & 75.15 & 68.51 & 95.94 & 86.93 & 70.50 & 83.46 \\
	UD + Perseus & 93.35 & 73.26 & 78.99 & 75.30 & 61.26 & 94.73 & 86.48 & 71.06 & \color{black} {\cellcolor[HTML]{EF553B}} 81.68 \\
	UD + PROIEL & \color{black} {\cellcolor[HTML]{EF553B}} 80.04 & 74.23 & 79.87 & \color{black} {\cellcolor[HTML]{EF553B}} 61.56 & \color{black} {\cellcolor[HTML]{B6E880}} 87.37 & \color{black} {\cellcolor[HTML]{EF553B}} 94.12 & \color{black} {\cellcolor[HTML]{EF553B}} 75.08 & 76.09 & 81.79 \\
	UD + UDante & 93.42 & \color{black} {\cellcolor[HTML]{EF553B}} 37.67 & 80.28 & 74.65 & 51.83 & 95.23 & 88.60 & 32.18 & 83.91 \\
	UD + eFontes & \color{black} {\cellcolor[HTML]{B6E880}} 94.97 & \color{black} {\cellcolor[HTML]{B6E880}} 87.67 & \color{black} {\cellcolor[HTML]{B6E880}} 83.17 & \color{black} {\cellcolor[HTML]{B6E880}} 79.61 & 78.60 & 96.35 & \color{black} {\cellcolor[HTML]{B6E880}} 96.20 & \color{black} {\cellcolor[HTML]{B6E880}} 81.45 & 86.82 \\
	\hline
\end{tabular}
\caption{The results of the various training scenarios described in Section \ref{sec:system}. Green color indicates the best result, while red indicates the worse result for a given subcorpus-task combination. The results involving eFontes data (\textit{baseline} and \textit{UD + eFontes}) are reported for the cross-validation scheme, i.e. the evaluated model was trained on the data excluding the testing subcorpus. For the other scenarios for each task only one model was trained and the results show its performance for the different subcorpora.}
\label{tab:full_split}
\end{table*}

The performance of the models for part of speech, morphological features, and lemma tagging tasks was assessed according to the scenarios described in Section \ref{sec:system}. 
The performance for each \textit{eFontes} sub-corpus was evaluated separately, as shown in Table \ref{tab:full_split}.

Overall, it is evident that the best results were obtained in the last scenario, where the model fine-tuned on the UD Latin data was further fine-tuned on specific \textit{eFontes} sub-corpora. 
In that scenario, lemmatization results ranged from 83.17 (Proceedings) to 96.35 (Science). 
Conversely, morphological features tagging showed varying outcomes, with accuracy rates of 78.60 for the Science sub-corpus and 87.67 for the Proceedings genre.
Regarding POS tagging, the system achieved the highest accuracy in the Annals genre (96.20) and lower accuracy in the Science sub-corpus (75.35). 
The character of the errors and possible reasons for the poor performance of the models in some of the sub-corpora are discussed in Section \ref{sec:qualitative}.

The comparison with the baseline models clearly demonstrates that it is possible to achieve moderate improvement (1-3 percentage points) through additional training on the UD data. The most signigicant difference is observed for the lemmatization task for the Science sub-corpus, with an improvement of almost 16 percentage points. 

When comparing the results between the scenarios referred to as \textit{UD all} and \textit{UD + UD corpus}, it is interesting to note that in most cases, using all the available data is not the optimal choice. Instead, for many task--sub-corpus combinations, a model further trained on a specific UD corpus yields better results. This suggests significant variability between the data in the collected corpora and indicates that more specific sub-corpus is better suited for achieving the best results.
It is also possible that combining specific UD corpus with \textit{eFontes} data could yield better results for some of the \textit{eFontes} sub-corpora. However, such a scenario would involve too many experimental combinations, so it was excluded from the setup.

In the UFeats task, the model fine-tuned on the PROIEL corpus achieved a performance score of 87.37, surpassing the model for the last scenario by more than 8 percentage points. For lemmatization, the model fine-tuned on all UD corpora showed only a minimal advantage of 0.10 percentage points.

In conclusion, the results highlight the significance of the availability of high-quality annotated corpora for improving the accuracy of models. While the advantages of training with \textit{eFontes} data may not be immediately obvious for some genres and tasks (e.g., lemmatization for Science and Annals genre), 
the difference is significant for the majority of them. In the next section, we discuss main results of qualitative error analysis for lemmatization and PoS tagging tasks.





\subsection{Qualitative Analysis}
\label{sec:qualitative}
\subsubsection{Lemmatization}
Following the example set by \citet{WROBELNOWAK}, we conducted an in-depth qualitative analysis to investigate the nature of tagging errors, identify their sources, and determine ways to improve the results.

Overall, the analysis revealed that a significant number of lemmatization and POS tagging errors could have been easily reduced by simplifying the task and harmonizing the training datasets. Trivial errors included, for example, frequent mislabeling of the SYM tokens: over 10\% of the total number of lemmatization errors were found to be due to the way the model handled mathematical notation in the Science sub-corpus, which explains its low accuracy (see Figure \ref{fig:lemmas_genre}).

In the gold corpus, tokens that represent elements of mathematical notation, such as points, lines, angles, and geometric shapes (e.g.,\textit{A} or \textit{CD}), as well as fractions (e.g., \textit{1amXI}), were uniformly labeled as SYM, with their lemma set to an underscore (\_). This decision was made to differentiate occurrences of symbolic tokens from those of "meaningful" words, particularly in instances where homonymy could occur, such as with the adposition \textit{ab}. Another practical reason for this approach was to offer a simplified lemma representation for tokens whose interpretation might not be immediately clear. However, our model would frequently missclassify such tokens as nouns and assigned them "meaningful" lemmas, such as \textit{resp.}, \emph{a}, or \emph{cd}.\footnote{In a somewhat similar manner, nearly 50 lemmatization errors resulted from the system replacing the editorial symbol \*\/\* with a plus sign (+).} In future versions of the model, it appears reasonable to consider restricting the lemmatization of symbolic units. For clarity, similar errors have been excluded from the discussion in the subsequent analysis.

\begin{figure}[htb]
\includegraphics[scale=0.11]{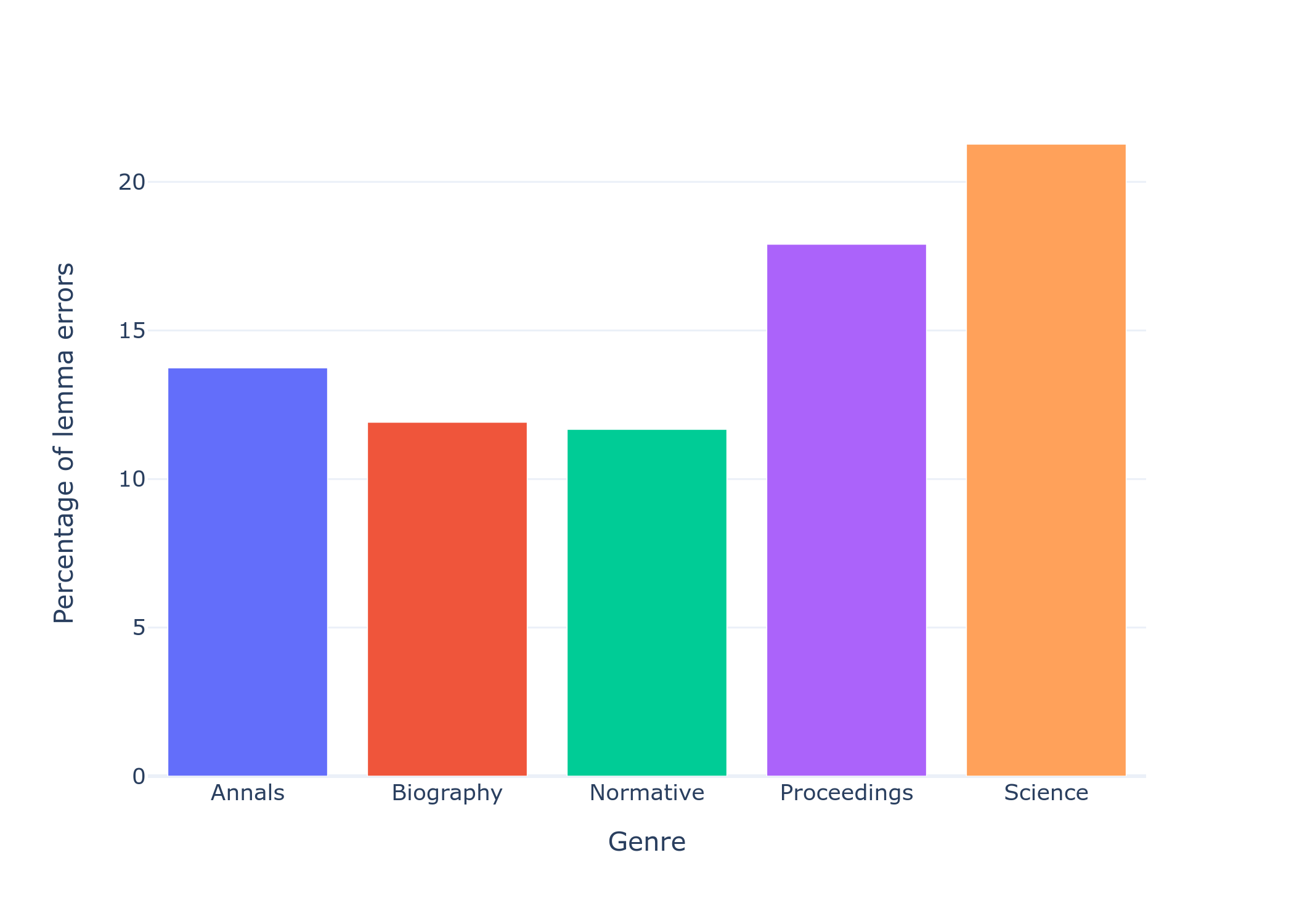}
\centering
\caption{Lemmatization task: genre distribution of errors.}
\end{figure}
\label{fig:lemmas_genre}

Beyond these trivial instances, further examination revealed that lemmatization errors primarily arose from the handling of orthographic variation in medieval Latin texts and a significant presence of Latinized Polish terms and proper names in the annotated texts, for which the training corpora lacked sufficient data.

\begin{table}[!htb]
\scriptsize
\centering
\begin{tabular}{lc|lc|lc}
\multicolumn{6}{c}{Position} \\
\multicolumn{2}{c}{initial} & \multicolumn{2}{c}{middle} & \multicolumn{2}{c}{final} \\
\toprule
pattern & count & pattern & count & pattern & count \\
\midrule
u:v & {\cellcolor[HTML]{023858}} \color[HTML]{F1F1F1} 289 & u:v & {\cellcolor[HTML]{023858}} \color[HTML]{F1F1F1} 281 & a:us & {\cellcolor[HTML]{023858}} \color[HTML]{F1F1F1} 107 \\
k:c & {\cellcolor[HTML]{BCC7E1}} \color[HTML]{000000} 95 & t:c & {\cellcolor[HTML]{034D79}} \color[HTML]{F1F1F1} 260 & z:us & {\cellcolor[HTML]{034D79}} \color[HTML]{F1F1F1} 100 \\

i:j & {\cellcolor[HTML]{D9D8EA}} \color[HTML]{000000} 66 & a: & {\cellcolor[HTML]{E9E5F1}} \color[HTML]{000000} 55 & s:m & {\cellcolor[HTML]{2F8BBE}} \color[HTML]{F1F1F1} 76 \\

h: & {\cellcolor[HTML]{F4EDF6}} \color[HTML]{000000} 18 & k:c & {\cellcolor[HTML]{EFE9F3}} \color[HTML]{000000} 47 & o:us & {\cellcolor[HTML]{89B1D4}} \color[HTML]{000000} 59 \\

a: & {\cellcolor[HTML]{FAF2F8}} \color[HTML]{000000} 17 & :h & {\cellcolor[HTML]{F9F2F8}} \color[HTML]{000000} 30 & m:s & {\cellcolor[HTML]{C0C9E2}} \color[HTML]{000000} 46 \\

\end{tabular}
\caption{5 most frequent lemma confusion patterns.}
\end{table}
\label{fig:confusion_patters}

The most common group of errors (see Table \ref{fig:confusion_patters}) stems from the misspelling of the Latin bilabial \emph{\/v\/}. While in the gold corpus it was rendered as \textit{u} for both consonant and vowel, the model would yield \textit{v} for the consonant variant, thus preferred \emph{uideo}, \emph{ciuitas}, and \emph{uiuo} would be replaced with \emph{video}, \emph{civitas}, and \emph{vivo}.

The second prevalent group of errors is related to the spelling of the Latin CV group \emph{-ti-} which would be often spelled as \emph{-ci-} in the medieval texts. To minimize spelling variance, in the \textit{eFontes} corpus the group is always represented as \textit{-ti-}. However, the model often opted to output \textit{-ci-}: instead of \emph{laurentius}, \emph{gratia}, or \emph{pretiosus}, it produced \emph{laurencius}, \emph{gracia}, or \emph{preciosus}.

Third, the models substituted the letter \emph{k} with \textit{c}, despite the former being standard in medieval documents. This error particularly affects the ``Proceedings'' and ``Biography'' sub-corpora where it occurs in the spelling of Polish proper names, such as \emph{kinga} $\rightarrow$ \emph{cinga}, \emph{thokarz} $\rightarrow$ \emph{thocarz}, and \emph{stassek} $\rightarrow$ \emph{stassec}.

The error types discussed so far account for nearly 40\% of the lemmatization errors produced by the system. Other notable categories include:
\begin{itemize}
\item substitution of the diphthongs \emph{ae} and \emph{oe} with \textit{e} (\emph{daemon} $\rightarrow$ \emph{demon}, \emph{aequidisto} $\rightarrow$ \emph{equidisto}, or \emph{aeuum} $\rightarrow$ \emph{evum}; and \emph{dioecesanus} $\rightarrow$ \emph{diocesanus}, \emph{uesperae} $\rightarrow$ \emph{vespera});
\item addition or omission of the \emph{h} consonant despite the standardized spelling adopted in the gold corpus, for example, \emph{tomco} $\rightarrow$ \emph{thomcus}, \emph{platea} $\rightarrow$ \emph{plathea}, \emph{iohannes} $\rightarrow$ \emph{joannes}, and \emph{hungaria} $\rightarrow$ \emph{ungaria}.
\end{itemize}

The preference for specific spelling cannot be always traced back to the form of the token but should rather be attributed to the structure of the training datasets, which were not normalized or harmonized beforehand. For example, concerning the \textit{u:v} alternation, the LLCT and UDante corpora exclusively use \textit{u} for lemma forms, whereas the Perseus and PROIEL corpora employ both \textit{v} and \textit{u} to distinguish between consonantal and vowel variations. In the ITTB corpus, the letter \textit{v} does not appear in either surface forms or lemmas.

Regarding the \textit{-ti-}:\textit{-ci-} alternation, it is observed exclusively in the LLCT corpus. Over 30 occurrences include proper names, such as \emph{Baruncio} \(\rightarrow\) \emph{Barontius}, \emph{Laurencii} \(\rightarrow\) \emph{Laurentius}), as well as common nouns, for example, \emph{presencia} \(\rightarrow\) \emph{praesentia}, \emph{palacio} \(\rightarrow\) \emph{palatium}. However, within the same corpus, these lemmas are more often spelled in a standardized manner, with forms like \emph{praesentia}, \emph{Laurentii}, and so forth.

The LLCT corpus, along with the \textit{eFontes} corpus, is unique in providing data on the \emph{k}:\emph{c} alternation. Its attestations, however, are limited to occurrences of a single word only, namely the noun \emph{karitas} : \emph{caritas}. \footnote{The corpus includes numerous examples of the \emph{k} spelling for both surface form and lemma, largely due to the prevalence of terms like \emph{kalendae}. Notably, it includes 5 examples of the spelling variation \emph{calendae} : \emph{kalendae}.}

The model's inability to correctly lemmatize words that, according to classical norm, should be spelled with a diphthong \textit{ae} or \textit{oe} is noteworthy, as the UD corpora seem to offer ample evidence for such normalization. Specifically, the LLCT corpus contains over 200 instances of the \textit{e}:\textit{oe} alternation, although the range of words concerned is limited primarily to forms of \textit{poena} and \textit{oboedientia}. Additionally, the corpus contains more than 1000 instances of the \textit{e}:\textit{ae} variance. Both types of alternation also appear, though less frequently, in the ITTB, UDante, and PROIEL corpora, but are virtually absent in the Perseus corpus.

The situation becomes even less clear regarding the spelling of the \textit{h} consonant for the UD corpora provide evidence of its usage in post-consonantal and intervocalic positions at the beginning and in the middle of a word.

\subsubsection{Part of Speech tagging}

\begin{figure}[tb]
\centering
\begin{center}
\includegraphics[scale=0.25]{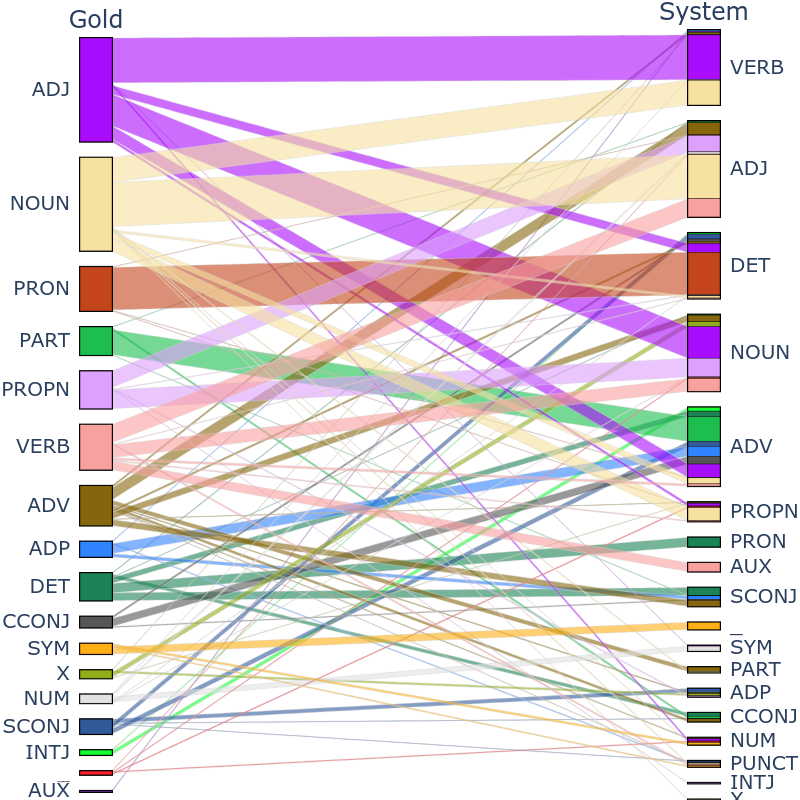} \caption{The UPOS task: break-down of error types.}
\label{2024_PoS_confusion_genre_flow}
\end{center}
\end{figure}

Setting aside the errors arising from the misinterpretation of SYM tokens, the parts of speech most frequently mislabeled were adjectives, nouns, verbs, pronouns, adverbs, proper nouns, particles, and determiners (Figure \ref{2024_PoS_confusion_genre_flow}). Many of these errors tend to recur in PoS tagging tasks, as demonstrated by \citet{WROBELNOWAK}. They can often be traced back to the derivational relationships between words and seem to arise as a result of insufficient context, which does not allow for choosing between multiple interpretations.

This phenomenon includes the frequent mislabeling of adjectives as participial forms of verbs. In the dataset analyzed, a common error involved the high-frequency phrase \emph{iudicium bannitum}, a technical term denoting a category of court trials in medieval Poland. Although the adjective \textit{bannitus} originates from the verb \textit{bannio}, its adjectival meaning has become fully lexicalized, leading to its classification as an adjective in Medieval Latin dictionaries. A similar situation occurs with the adjective \emph{aequidistans} 'equidistant'. The verb \emph{aequidisto}, to which it is linked, is even less frequently attested in medieval data than the verb \textit{bannio} discussed above.

Nouns were most frequently misidentified as adjectives, accounting for 47\% of such errors in the analyzed data. This error impacted both deadjectival nouns attested already in Classical Latin, such as \emph{sanctus} 'a saint' and \emph{bonum} 'good behaviour, deed etc.', and medieval terms like \emph{grossus} 'type of currency', which are not found in the UD corpora but as adjectives.

In the reverse scenario of mislabeling adjectives and nouns, participial forms of verbs, such as \textit{debitus} or \textit{contractus}, are incorrectly annotated as adjectives (accounting for 40\% of errors) or as nouns (28\% of errors).



	\section{Conclusions}
	\label{sec:conclusions}
	The models presented in this study were designed for the automatic annotation of medieval texts across a wide range of domains, varying levels of formality, genres, and communicative contexts. Although they have achieved satisfactory results in most tasks, further research is certainly needed to conduct a rigorous comparison with existing solutions.

An examination of the tagging errors revealed that a considerable portion of them were minor in nature, suggesting they could be easily remedied in future versions of the system by simplifying the annotation of symbol units, such as fractions and editorial marks, for example. Other errors, in turn, stemmed from inconsistent spelling of Latin words in the training data or from insufficient evidence for preferred normalized spelling in the \textit{eFontes} corpus.

The study further indicated that training on manually annotated corpora, like \textit{eFontes}, considerably improves the accuracy of tagging where significant domain- or genre-specific variation of data may be observed.

Future research will focus on the challenges of benchmarking against existing systems, including GPT models, which are presently viewed as competitors to the custom models discussed in this paper. Additionally, plans include expanding the datasets to cover historically significant, yet unexplored, medieval genres such as poetry or medieval Latin documents.

Finally, the authors are currently working on an automated solution for tagging named entities in medieval Latin texts from Polish sources. A high-performing NER tagger, while significant in its own right, should help in mitigating some of the issues associated with processing vernacular proper names discussed in this paper.

	\section{Acknowledgments}
	This work was supported by the project \textit{eFontes. The Electronic Corpus of Polish Medieval Latin} (11H 17 0116 85) funded by the Polish Ministry of Science and by the grant of the PLGrid Infrastructure.
	
	
	\bibliographystyle{acl_natbib.sty}
	\bibliography{arxiv}

\begin{thebibliography}{21}
\providecommand{\natexlab}[1]{#1}

\bibitem[{Bon(2011)}]{bonOMNIAOutilsMethodes2011}
Bruno Bon. 2011.
\newblock \href {https://doi.org/10.4000/cem.12015} {Omnia : outils et
  méthodes numériques pour l’interrogation et l’analyse des textes
  médiolatins (3)}.
\newblock \emph{BUCEMA. Bulletin du centre d’études médiévales
  d’Auxerre}, (15):251--252.

\bibitem[{Camps et~al.(2021)Camps, Clérice, Duval, Ing, Kanaoka, and
  Pinche}]{campsCorpusModelsLemmatisation2021}
Jean-Baptiste Camps, Thibault Clérice, Frédéric Duval, Lucence Ing, Naomi
  Kanaoka, and Ariane Pinche. 2021.
\newblock \href {https://arxiv.org/abs/2109.11442} {Corpus and {{Models}} for
  {{Lemmatisation}} and {{POS-tagging}} of {{Old French}}}.
\newblock \emph{Preprint}, arxiv:2109.11442.

\bibitem[{Cecchini et~al.(2020{\natexlab{a}})Cecchini, Sprugnoli, Moretti, and
  Passarotti}]{cecchiniUDanteFirstSteps2020}
Flavio~M. Cecchini, Rachele Sprugnoli, Giovanni Moretti, and Marco Passarotti.
  2020{\natexlab{a}}.
\newblock \href {https://doi.org/10.4000/books.aaccademia.8653} {{{UDante}}:
  {{First Steps Towards}} the {{Universal Dependencies Treebank}} of
  {{Dante}}’s {{Latin Works}}}.
\newblock In Felice Dell'Orletta, Johanna Monti, and Fabio Tamburini, editors,
  \emph{Proceedings of the {{Seventh Italian Conference}} on {{Computational
  Linguistics CLiC-it}} 2020 : {{Bologna}}, {{Italy}}, {{March}} 1-3, 2021},
  Collana Dell'{{Associazione Italiana}} Di {{Linguistica Computazionale}},
  pages 99--105. {Accademia University Press}.

\bibitem[{Cecchini et~al.(2020{\natexlab{b}})Cecchini, Korkiakangas, and
  Passarotti}]{cecchini-etal-2020-new}
Flavio~Massimiliano Cecchini, Timo Korkiakangas, and Marco Passarotti.
  2020{\natexlab{b}}.
\newblock \href {https://aclanthology.org/2020.lrec-1.117} {A new {L}atin
  treebank for {U}niversal {D}ependencies: Charters between {A}ncient {L}atin
  and {R}omance languages}.
\newblock In \emph{Proceedings of the Twelfth Language Resources and Evaluation
  Conference}, pages 933--942, Marseille, France. European Language Resources
  Association.

\bibitem[{Cecchini et~al.(2018)Cecchini, Passarotti, Marongiu, and
  Zeman}]{lait-ud-2018}
Flavio~Massimiliano Cecchini, Marco Passarotti, Paola Marongiu, and Daniel
  Zeman. 2018.
\newblock Challenges in converting the \emph{Index Thomisticus} treebank into
  universal dependencies.
\newblock \emph{Proceedings of the Universal Dependencies Workshop 2018 (UDW
  2018)}.

\bibitem[{Eger et~al.(2016)Eger, Gleim, and
  Mehler}]{eger-etal-2016-lemmatization}
Steffen Eger, R{\"u}diger Gleim, and Alexander Mehler. 2016.
\newblock Lemmatization and {{Morphological Tagging}} in {{German}} and
  {{Latin}}: {{A Comparison}} and a {{Survey}} of the {{State-of-the-art}}.
\newblock In \emph{Proceedings of the Tenth International Conference on
  Language Resources and Evaluation ({{LREC}}'16)}, pages 1507--1513,
  {Portoro{\v z}, Slovenia}. {European Language Resources Association (ELRA)}.

\bibitem[{Goyens and Verbeke(2003)}]{goyensDawnWrittenVernacular2003}
Michèle Goyens and Werner Verbeke, editors. 2003.
\newblock \emph{The dawn of the written vernacular in Western Europe}.
\newblock Leuven University Press.

\bibitem[{Haug and J{\o}hndal(2008)}]{Haug2008CreatingAP}
Dag Trygve~Truslew Haug and Marius~L. J{\o}hndal. 2008.
\newblock \href {https://api.semanticscholar.org/CorpusID:204978005} {Creating
  a parallel treebank of the old indo-european bible translations}.
\newblock In \emph{Proceedings of the Second Workshop on Language Technology
  for Cultural Heritage Data (LaTeCH 2008)}, pages 27--34.

\bibitem[{Johnson et~al.(2021)Johnson, Burns, Stewart, Cook, Besnier, and
  Mattingly}]{johnson-etal-2021-classical}
Kyle~P. Johnson, Patrick~J. Burns, John Stewart, Todd Cook, Cl{\'e}ment
  Besnier, and William J.~B. Mattingly. 2021.
\newblock \href {https://doi.org/10.18653/v1/2021.acl-demo.3} {The {C}lassical
  {L}anguage {T}oolkit: {A}n {NLP} framework for pre-modern languages}.
\newblock In \emph{Proceedings of the 59th Annual Meeting of the Association
  for Computational Linguistics and the 11th International Joint Conference on
  Natural Language Processing: System Demonstrations}, pages 20--29, Online.
  Association for Computational Linguistics.

\bibitem[{Kestemont and
  De~Gussem(2016)}]{kestemontIntegratedSequenceTagging2016}
Mike Kestemont and Jeroen De~Gussem. 2016.
\newblock \href {https://arxiv.org/abs/1603.01597} {Integrated {{Sequence
  Tagging}} for {{Medieval Latin Using Deep Representation Learning}}}.
\newblock \emph{arXiv:1603.01597 [cs, stat]}.

\bibitem[{Riemenschneider and
  Frank(2023)}]{riemenschneiderExploringLargeLanguage2023}
Frederick Riemenschneider and Anette Frank. 2023.
\newblock \href {https://arxiv.org/abs/2305.13698} {Exploring {{Large Language
  Models}} for {{Classical Philology}}}.
\newblock \emph{Preprint}, arxiv:2305.13698.

\bibitem[{Sommerschield et~al.(2023)Sommerschield, Assael, Pavlopoulos,
  Stefanak, Senior, Dyer, Bodel, Prag, Androutsopoulos, and {de
  Freitas}}]{sommerschieldMachineLearningAncient2023}
Thea Sommerschield, Yannis Assael, John Pavlopoulos, Vanessa Stefanak, Andrew
  Senior, Chris Dyer, John Bodel, Jonathan Prag, Ion Androutsopoulos, and Nando
  {de Freitas}. 2023.
\newblock \href {https://doi.org/10.1162/coli_a_00481} {Machine {{Learning}}
  for {{Ancient Languages}}: {{A Survey}}}.
\newblock \emph{Computational Linguistics}, 49(3):703--747.

\bibitem[{Sprugnoli et~al.(2022)Sprugnoli, Passarotti, Cecchini, Fantoli, and
  Moretti}]{sprugnoliOverviewEvaLatin20222022}
Rachele Sprugnoli, Marco Passarotti, Flavio~Massimiliano Cecchini, Margherita
  Fantoli, and Giovanni Moretti. 2022.
\newblock \href {https://aclanthology.org/2022.lt4hala-1.29} {Overview of the
  {{EvaLatin}} 2022 {{Evaluation Campaign}}}.
\newblock In \emph{Proceedings of the {{Second Workshop}} on {{Language
  Technologies}} for {{Historical}} and {{Ancient Languages}}}, pages 183--188.
  {European Language Resources Association}.

\bibitem[{Sprugnoli et~al.(2020)Sprugnoli, Passarotti, Cecchini, and
  Pellegrini}]{sprugnoliOverviewEvaLatin20202020}
Rachele Sprugnoli, Marco Passarotti, Flavio~Massimiliano Cecchini, and Matteo
  Pellegrini. 2020.
\newblock \href {https://aclanthology.org/2020.lt4hala-1.16} {Overview of the
  {{EvaLatin}} 2020 {{Evaluation Campaign}}}.
\newblock In \emph{Proceedings of {{LT4HALA}} 2020 - 1st {{Workshop}} on
  {{Language Technologies}} for {{Historical}} and {{Ancient Languages}}},
  pages 105--110. {European Language Resources Association (ELRA)}.

\bibitem[{Stotz(1996-2004)}]{stotzHandbuchZurLateinischen1996a}
Peter Stotz. 1996-2004.
\newblock \emph{Handbuch zur lateinischen Sprache des Mittelalters}, volume
  1--5.
\newblock C. H. Beck.

\bibitem[{Van~Nguyen et~al.(2021)Van~Nguyen, Lai, Veyseh, and
  Nguyen}]{van2021trankit}
Minh Van~Nguyen, Viet~Dac Lai, Amir Pouran~Ben Veyseh, and Thien~Huu Nguyen.
  2021.
\newblock Trankit: A light-weight transformer-based toolkit for multilingual
  natural language processing.
\newblock \emph{arXiv preprint arXiv:2101.03289}.

\bibitem[{Wolf et~al.(2020)Wolf, Debut, Sanh, Chaumond, Delangue, Moi, Cistac,
  Rault, Louf, Funtowicz, Davison, Shleifer, von Platen, Ma, Jernite, Plu, Xu,
  Le~Scao, Gugger, Drame, Lhoest, and Rush}]{wolf2020transformers}
Thomas Wolf, Lysandre Debut, Victor Sanh, Julien Chaumond, Clement Delangue,
  Anthony Moi, Pierric Cistac, Tim Rault, Remi Louf, Morgan Funtowicz, Joe
  Davison, Sam Shleifer, Patrick von Platen, Clara Ma, Yacine Jernite, Julien
  Plu, Canwen Xu, Teven Le~Scao, Sylvain Gugger, Mariama Drame, Quentin Lhoest,
  and Alexander Rush. 2020.
\newblock \href {https://doi.org/10.18653/v1/2020.emnlp-demos.6} {Transformers:
  State-of-the-art natural language processing}.
\newblock In \emph{Proceedings of the 2020 Conference on Empirical Methods in
  Natural Language Processing: System Demonstrations}, pages 38--45, Online.
  Association for Computational Linguistics.

\bibitem[{Wr{\'o}bel and Nowak(2022)}]{WROBELNOWAK}
Krzysztof Wr{\'o}bel and Krzysztof Nowak. 2022.
\newblock \href {https://aclanthology.org/2022.lt4hala-1.31} {Transformer-based
  part-of-speech tagging and lemmatization for {L}atin}.
\newblock In \emph{Proceedings of the Second Workshop on Language Technologies
  for Historical and Ancient Languages}, pages 193--197, Marseille, France.
  European Language Resources Association.

\bibitem[{Wróbel(2020)}]{publ252933}
Krzysztof Wróbel. 2020.
\newblock \href {http://poleval.pl/files/poleval2020.pdf} {Kftt : Polish full
  neural morphosyntactic tagger}.
\newblock In Maciej Ogrodniczuk and Łukasz Kobyliński, editors,
  \emph{Proceedings of the PolEval 2020 Workshop}, pages 47--53. Institute of
  Computer Sciences, Polish Academy of Sciences, Warszawa.

\bibitem[{Xue et~al.(2022)Xue, Barua, Constant, Al-Rfou, Narang, Kale, Roberts,
  and Raffel}]{10.1162/tacl_a_00461}
Linting Xue, Aditya Barua, Noah Constant, Rami Al-Rfou, Sharan Narang, Mihir
  Kale, Adam Roberts, and Colin Raffel. 2022.
\newblock \href {https://doi.org/10.1162/tacl_a_00461} {{ByT5: Towards a
  Token-Free Future with Pre-trained Byte-to-Byte Models}}.
\newblock \emph{Transactions of the Association for Computational Linguistics},
  10:291--306.

\bibitem[{Xue et~al.(2021)Xue, Constant, Roberts, Kale, Al-Rfou, Siddhant,
  Barua, and Raffel}]{xue-etal-2021-mt5}
Linting Xue, Noah Constant, Adam Roberts, Mihir Kale, Rami Al-Rfou, Aditya
  Siddhant, Aditya Barua, and Colin Raffel. 2021.
\newblock \href {https://doi.org/10.18653/v1/2021.naacl-main.41} {m{T}5: A
  massively multilingual pre-trained text-to-text transformer}.
\newblock In \emph{Proceedings of the 2021 Conference of the North American
  Chapter of the Association for Computational Linguistics: Human Language
  Technologies}, pages 483--498, Online. Association for Computational
  Linguistics.

\end{thebibliography}
	
	
\end{document}